\documentclass[sigconf]{acmart}
\AtBeginDocument{%
  }

\usepackage{algorithm}
\usepackage{algpseudocode}
\usepackage{amsmath}
\usepackage{bm}
\newcommand{\vect}[1]{\bm{#1}}
\usepackage{hyperref}
\usepackage{cleveref}

\usepackage{array}
\newcolumntype{C}[1]{>{\centering\arraybackslash}p{#1}}
\usepackage{multicol}
\usepackage{multirow}
\usepackage{graphicx}
\usepackage{booktabs}
\usepackage{xcolor}
\usepackage{ulem}
\usepackage{subcaption}
\begin{document}

\title{Agentic Spatio-Temporal Grounding via Collaborative Reasoning}


% \author{Anonymous}
% % \authornote{Both authors contributed equally to this research.}
% % \orcid{1234-5678-9012}
% \affiliation{%
%   \institution{Anonymous Institute}
%   \country{Anonymous Country}
% }
% \email{Anonymous email}

\author{Heng Zhao}
\affiliation{%
  \institution{CFAR, IHPC, Agency for Science, Technology and Research(A*STAR)}
  \country{Singapore}
}
\affiliation{%
  \institution{CCDS, Nanyang Technological University}
  \country{Singapore}
}
% \email{zhaoh@a-star.edu.sg}
  
\author{Yew-Soon Ong}
\affiliation{%
  \institution{CFAR, IHPC, Agency for Science, Technology and Research(A*STAR)}
  \country{Singapore}
}
\affiliation{%
  \institution{CCDS, Nanyang Technological University}
  \country{Singapore}
}
% \email{asysong@ntu.edu.sg}

\author{Joey Tianyi Zhou}
\affiliation{%
  \institution{CFAR, IHPC, Agency for Science, Technology and Research(A*STAR)}
  \country{Singapore}
}
% \email{joey_zhou@a-star.edu.sg}

%% the author to define a more concise list\usepackage{bm}



% joey comments
% 1. Existing methods are inefficient in both training and inference due to the redundancy of the spatial localization performed on every frame within the predicted time span. This also leads to dense annotation requirements and redundant spatial reasoning for supervised approaches where the trade-off between performance and generalization ability is also a concern. Weakly-supervised methods alleviate the situation with pre-processed tubes but are still approximating the dataset-level train-and-fit paradigm with an inferior performance.
% -> reduce
% 2. simplify fig1:d into agentic abstraction but emphsize comparative advantage
% 3. figure 2 cap: briefly include all symbols used

\begin{abstract}
Spatio-Temporal Video Grounding (STVG) aims to retrieve the spatio-temporal tube of a target object or person in a video given a text query. Most existing approaches perform frame-wise spatial localization within a predicted temporal span, resulting in redundant computation, heavy supervision requirements, and limited generalization. Weakly-supervised variants mitigate annotation costs but remain constrained by the dataset-level train-and-fit paradigm with an inferior performance. To address these challenges, we propose the Agentic Spatio-Temporal Grounder (ASTG) framework for the task of STVG towards an open-world and training-free scenario. Specifically, two specialized agents SRA (Spatial Reasoning Agent) and TRA (Temporal Reasoning Agent) constructed leveraging on modern Multimoal Large Language Models (MLLMs) work collaboratively to retrieve the target tube in an autonomous and self-guided manner. Following a propose-and-evaluation paradigm, ASTG duly decouples spatio-temporal reasoning and automates the tube extraction, verification and temporal localization processes. With a dedicate visual memory and dialogue context, the retrieval efficiency is significantly enhanced. Experiments on popular benchmarks demonstrate the superiority of the proposed approach where it outperforms existing weakly-supervised and zero-shot approaches by a margin and is comparable to some of the fully-supervised methods.
\end{abstract}

% 7:    825566909
% 77:   818579280
% 7777: 825584744
% 1. grounded visual cues affect/improve hallucination?
% 3. semantic entropy(last layer) / llm uncertainty (middle layers research?)


%    <concept>
%        <concept_id>10010147.10010178.10010224.10010225.10010227</concept_id>
%        <concept_desc>Computing methodologies~Scene understanding</concept_desc>
%        <concept_significance>500</concept_significance>
%        </concept>
%  </ccs2012>
% \ccsdesc[500]{Computing methodologies~Scene understanding}

% \keywords{AI Agent, Spatio-Temporal Video Grounding, Multi-modal Large Language Models, Reasoning, Training-Free}



\settopmatter{printacmref=false}
\renewcommand\footnotetextcopyrightpermission[1]{}
\acmConference{}{}{}

\maketitle

% \begin{figure}[!t]
%   \centering

%   % ---- Left column ----
%   \begin{minipage}[t]{0.44\linewidth}
%     \vspace{0pt}
%     \centering
%     \subcaptionbox{Spatial Reasoning\label{fig:1a}}{
%       \includegraphics[width=\linewidth]{figs/i1.pdf}
%     }\par\vspace{1pt}

%     % \vspace{4pt}

%     \subcaptionbox{Joint Reasoning\label{fig:1c}}{
%       \includegraphics[width=\linewidth]{figs/i3.pdf}}
    
%   \end{minipage}
%   \hfill
%   % ---- Right column ----
%   \begin{minipage}[t]{0.55\linewidth}
%     \vspace{0pt}
%     \centering
    
%     % \vspace{4pt}

%     \subcaptionbox{Temporal Reasoning\label{fig:1b}}{\includegraphics[width=\linewidth]{figs/i2.pdf}
%     }\par\vspace{1pt}
    
%     \subcaptionbox{ASTG (Ours)\label{fig:1d}}{
%       \includegraphics[width=\linewidth]{figs/i4_new.pdf}}
    
%   \end{minipage}

%   \caption{
%   (a): Static spatial reasoning ignores dynamic temporal context. (b): Temporal reasoning with holistic video input cannot infer spatial information in a dense manner. (c): Joint-reasoning systems repeat spatial reasoning process on each of the frames which is often highly redundant. (d): Unlike prior joint reasoning methods that re-infer spatial grounding independently across frames, ASTG enables collaborative hypothesis-driven reasoning between SRA and TRA, where spatial candidates are proposed once and iteratively verified temporally. }
%   \label{fig:motivation}
% \end{figure}

\begingroup
\captionsetup[subfigure]{skip=1pt}

% your figure code here

\begin{figure}[!t]
  \centering

  % ================= Row 1 =================
  % ---- Left: (a) over (b) ----
  \begin{minipage}[t]{0.48\linewidth}
    \vspace{0pt}
    \centering
    
    \subcaptionbox{Spatial Reasoning\label{fig:1a}}{%
      \includegraphics[width=\linewidth]{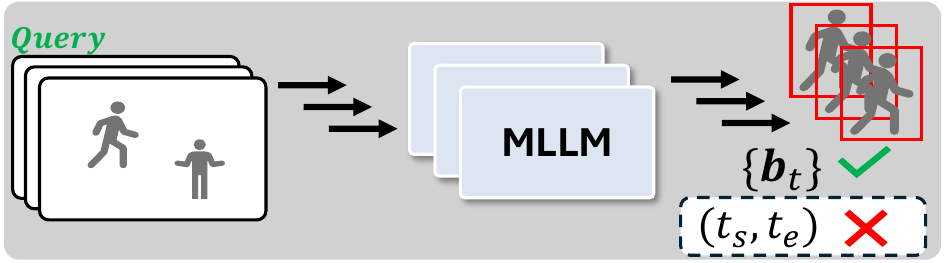}%
    }\par\vspace{2pt}

    % \vspace{-0.2em}
    \subcaptionbox{Temporal Reasoning\label{fig:1b}}{%
      \includegraphics[width=\linewidth]{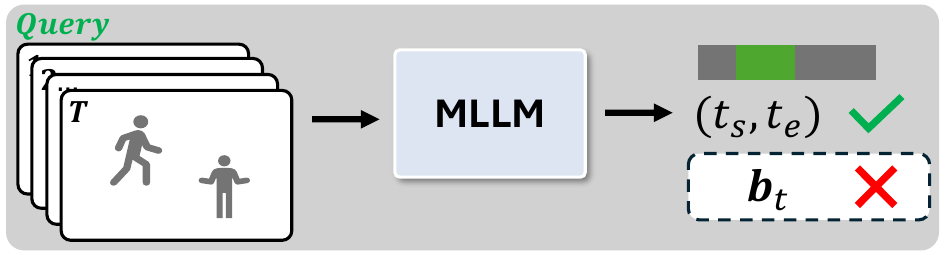}%
    }
  \end{minipage}
  \hfill
  % ---- Right: (c) ----
  \begin{minipage}[t]{0.48\linewidth}
    \vspace{0pt}
    \centering
    \subcaptionbox{Joint Reasoning\label{fig:1c}}{%
      \includegraphics[width=\linewidth]{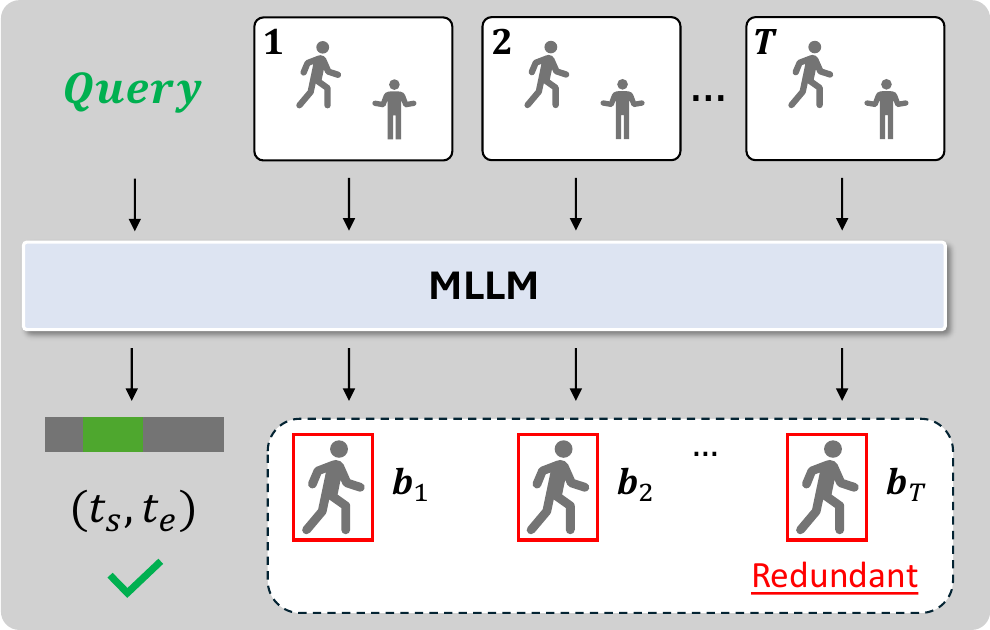}%
    }
  \end{minipage}

  \par\vspace{4pt}

  % ================= Row 2 =================
  % ---- Full width: (d) ----
  \begin{minipage}[t]{0.95\linewidth}
    \centering
    \subcaptionbox{ASTG (Ours)\label{fig:1d}}{%
      \includegraphics[width=\linewidth]{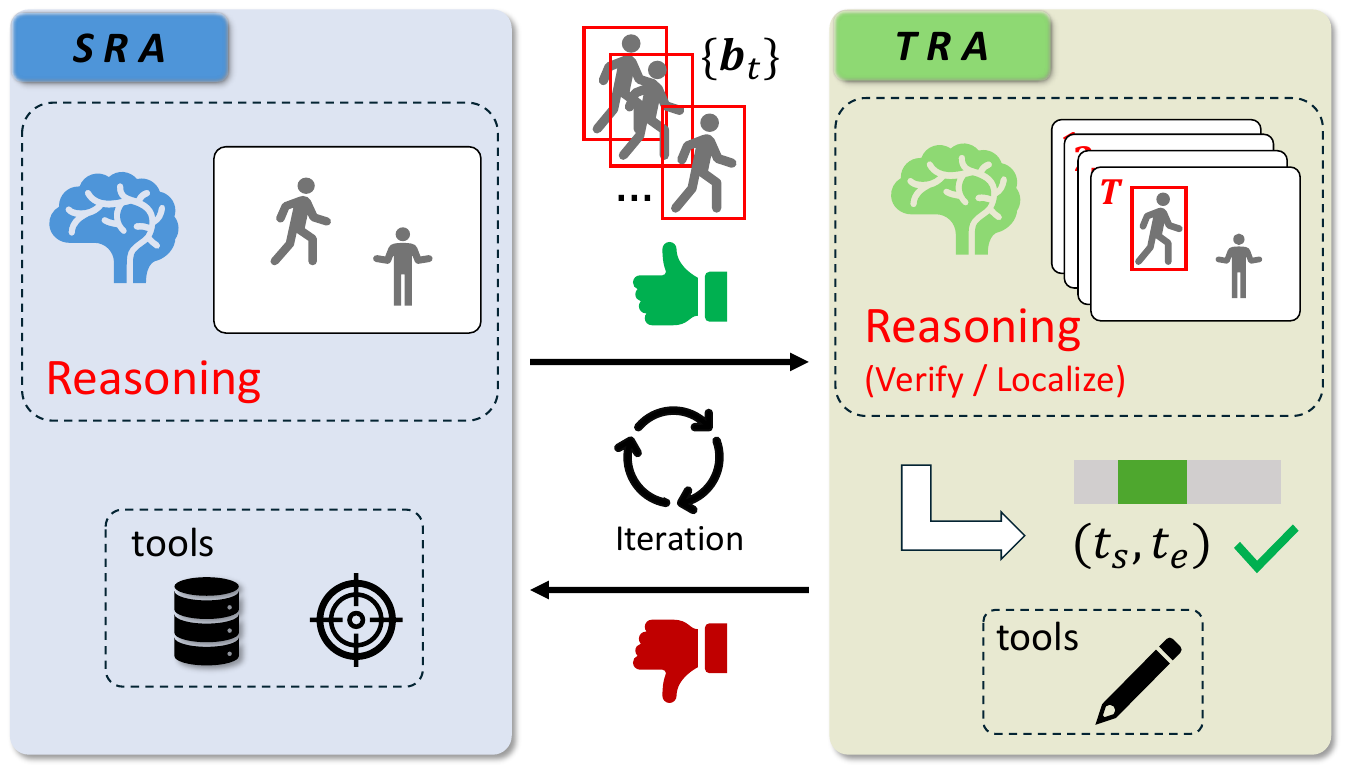}%
    }
  \end{minipage}

  \caption{%
  (a) Static spatial reasoning ignores dynamic temporal context.
  (b) Temporal reasoning with holistic video input cannot infer spatial information in a dense manner.
  (c) Joint-reasoning systems repeat spatial reasoning on each frame, leading to redundancy.
  (d) ASTG formulates STVG as a collaborative agentic process, where spatial candidates are proposed once and iteratively verified by Spatial and Temporal Reasoning Agents (SRA and TRA); leveraging tools such as candidate memory, object tracker and visual marker.
  }
  \label{fig:motivation}
\end{figure}

\endgroup

\section{Introduction}
Spatio-Temporal Video Grounding (STVG) is to locate a target object both spatially and temporally according to a query text which usually contains static visual cues and dynamic action cues. The task is highly relevant to real-world applications such as video surveillance, security and forensic analysis~\cite{surveil1, surveil2}, video editing, media production and content creation~\cite{videdit1,videdit2}, human-robot interaction and embodied AI~\cite{embodied1, embodied2, embodied3}, video search, retrieval and knowledge discovery~\cite{vidunderstanding1, vidunderstanding2} and so on. 

Unlike static image data, video data exhibit high dimensionality in both space and time, where spatially there may be multiple target entities for disambiguation and temporally all visual contents may evolve continuously at different pace. This poses a dilemma for grounding systems: performing spatial reasoning on each frame individually ignores the temporal cues (~\Cref{fig:1a}); analyzing the video as a whole cannot yield accurate spatial coordinates (~\Cref{fig:1b}); and the repeated spatial reasonings for each of the frames in joint-reasoning systems (~\Cref{fig:1c}) are highly redundant because the spatial displacement of the same object between adjacent frames is limited. In addition, these systems require dense spatial and temporal annotations to train. The scarce of such data and learning paradigm limit both their generalizability and efficiency in real-world applications.

To tackle these challenges, we propose Agentic Spatio-Temporal Grounder (ASTG): a training-free framework for spatio-temporal grounding in the wild; where the spatial reasoning and temporal reasoning process are duly decoupled and assigned to two specialized agents who work together in a collaborative manner to achieve the joint-reasoning following a propose-and-evaluate paradigm. Concretely, as illustrated in~\Cref{fig:1d}, the Spatial Reasoning Agent (SRA) provides the spatial coordinates of the proposed candidate via spatial reasoning only once on a certain frame; by invoking tracking tools such as SAM~\cite{sam2}, the candidate's spatio-temporal tube masks is retrieved and sent to TRA for verification. TRA invokes visual marking tools to apply the tube masks as spatial prompts to the frames and then perform temporal reasoning to validate the candidate against the given sentence query. Once verified, TRA will further trim it temporally via additional reasoning. In case of rejection, TRA will initiate negotiations with SRA iteratively in a collaborative manner to retrieve the correct candidate tube. Additionally, we propose a visual memory for SRA to filter out the candidate tubes that had been verified, effectively reducing the workload for TRA and the redundant reasoning steps. We also maintain a dialogue context between the specialized agents to better guide and adjust the responses of the SRA in an adaptive and autonomous manner. Experiments conducted on three widely adopted datasets demonstrate the superiority of our proposed framework over existing training-efficient approaches including weakly-supervised and zero-shot methods. Notably, the performance of the proposed method is already comparable to some of the fully-supervised methods. We summarize our contributions as follows:
\begin{itemize}
    \item We propose \textbf{Agentic Spatio-Temporal Grounder (ASTG)} for zero-shot and training-free STVG, an agentic framework with two agents working collaboratively to retrieve the fine-grained target candidate tube masks and bounding boxes. To the best of our knowledge, this is the first exploration of agent design for the task of STVG under a zero-shot paradigm.
    \item We introduce a \textbf{visual memory} and \textbf{dialogue context} to the agentic framework, which enables the autonomous joint-reasoning process and also adds transparency to the reasoning process where the intermediate results on the candidates can also serve as additional information for video analytics.
    \item Extensive experiments on multiple STVG benchmarks demonstrate that ASTG achieves state-of-the-art performance among training-efficient methods, while exhibiting strong generalization to open-vocabulary queries and unconstrained videos, suggesting that agentic, tool-augmented reasoning offers a promising direction beyond fully-supervised STVG models.
\end{itemize}

\section{Related Workds}

\noindent\textbf{Visual Grounding}. Grounding natural language in multi-media has been well established via image-level visual grounding~\cite{refcoco, refcocog, mattnet, faoa, w2p} which only handles static spatial reasoning. However, extending visual grounding from images to videos introduces additional challenges due to the temporal dimension, objects may undergo significant appearance changes caused by motion blur, occlusion, scale variation, or viewpoint shifts, while actions and interactions are often defined by subtle temporal dependencies rather than isolated frames. Early research~\cite{JHMDB, ucf101-24, ava} formulates the atomic action detection as an extension of object detection and action classification with pre-trained video feature extractors~\cite{3dcnn, c3d} and fixed action label sets. Later studies focus on video moment retrieval~\cite{vmr1,vmr2,vmr3,vmr4} which extends the query into free form texts but the actions and temporal cues are usually implicitly assumed on a single target for relatively simple scenes, eliminating the need for spatial disambiguation. In contrast, STVG~\cite{stgrn, hcstvg} requires both spatial and temporal disambiguation due to the presence of multiple candidate objects and actions, where models need to output both the temporal boundary and the target's spatial coordinates on each of the frames. Fully-supervised methods~\cite{stgrn, stvgbert, tubedetr,stcat,csdvls, cgstvg} achieve superior performance but require massive dense annotated training data and suffer from generalization issue; while weakly-supervised methods~\cite{awgu,winner,vcma,cospal,stpro} alleviate the requirement but lags in performance.

\noindent\textbf{Zero-shot STVG}.
Recently, researchers~\cite{e3m, lnstvg, realvg} try to tackle STVG in a zero-shot or training-free manner with the rise of foundation models and MLLMs. E3M~\cite{e3m} proposed an Expectation Maximization framework for zero-shot STVG leveraging the foundation model CLIP and pre-trained object detectors for the tube extraction. Although being training-free, the whole pipeline follows a similar procedure compared to some of the prompt based weakly-supervised methods~\cite{cvtp}. LN-STVG~\cite{lnstvg} exploits the text-to-image attention at token-level in trained MLLMs to find the best token for spatial grounding in videos without relying on bounding box supervision. The experiment indicates the best tokens are usually semantically meaningless such as \textit{``\_A"} and \textit{``:"}. Despite the findings, such training-free approach is not reasoning based and on the contrary, the strange behavior of MLLMs is largely unexplainable. Another recent work RealVG~\cite{realvg} constructs a fixed pattern framework with MLLM where the sentence query is parsed and rephrased as sub-question-answer pairs, which are later used for temporal relevance score calculation with MLLM in a multi-frame question-answering form over all frames in a heavily segmented manner. Despite the progress, the MLLMs in their frameworks are passively engaged where only limited or no explicit reasoning is needed; this results in reduced flexibility and adaptability and also largely restricted the potential of the MLLMs in an open-world scenario.

\begin{figure*}[t]
  \centering
  \includegraphics[width=\textwidth]{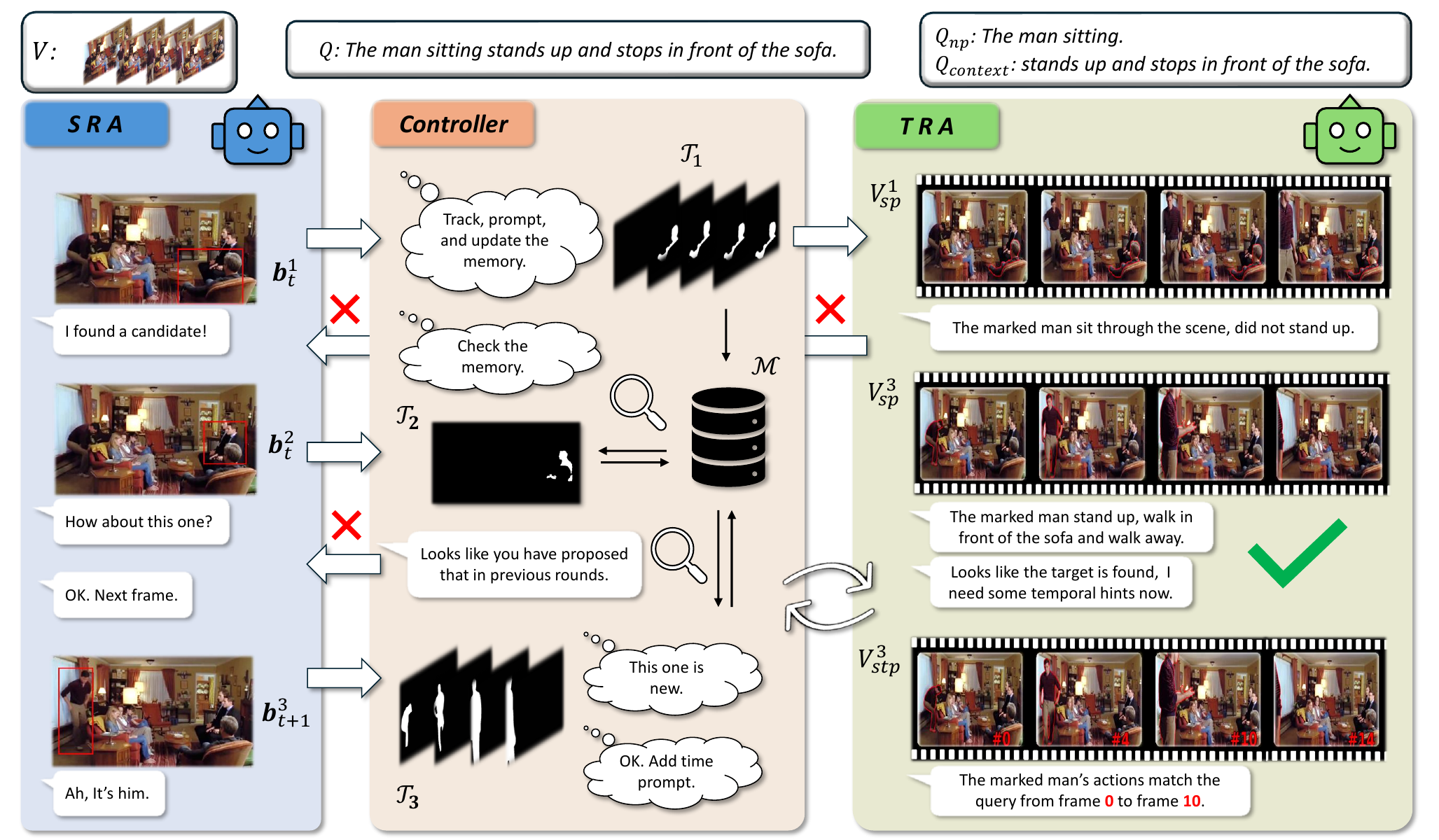}
  \caption{We propose the first Agentic Spatio-Temporal Grounding (ASTG) framework. Two agent modules Spatial Reasoning Agent (SRA) and Temporal Reasoning Agent (TRA) work in a collaborative manner to search and retrieve the correct target candidate tube $\mathcal{T}$. Spatial (red mask outlines) and temporal prompts (frames index, e.g. \texttt{"\#10"} ) are applied visually on the frames ($V_{sp}$ and $V_{stp}$) to assist the agents in the reasoning process. A visual memory module $\mathcal{M}$ and a dialogue context module are also proposed to improve the retrieval efficiency and guide the agents' behavior adaptively.}
  \label{fig:archi}
\end{figure*}

\section{Methodology}

In this section we elaborate the proposed agentic framework for spatio-temporla video grounding, where the Spatial and Temporal Agents work seamlessly by communicating to each other to retrieve and confirm the candidate tube that best matches the input query.

\subsection{Framework Overview}
Given a set of frames $V=\{f_1,\dots,f_T\}$ sampled from a video clip and a text description $Q$, the task of STVG is to retrieve a spatio-temporal tube consisting of a set of bounding boxes $\mathcal{B} = \{\bm{b}_{st}, ..., \bm{b}_{ed}\}$ where each of the bounding box $\bm{b}_i$ spatially grounds the target entity described by the input query $Q$ in corresponding frame $i$; and the starting and ending frame index $st$ and $ed$ also need to be inferred from the video $V$ and query $Q$.

%The intuition is that spatial and temporal reasoning should be decoupled duely because spatial reasoning is redundant if performed on all frames.
%As the video frames and the target spatial locations are contiguous, in intro
The challenge of solving this task under a training-free zero-shot paradigm is how to handle the dense spatial prediction on all frames while at the same time reasoning with temporal awareness in an accurate and efficient way. An apparent observation is that for any target entity in a video, its spatial locations on adjacent frames are contiguous. This implies spatial reasoning is redundant if performed on all frames. To reduce the need of such redundancy and maximize both spatial and temporal reasoning capability of the MLLM, we propose ASTG: a collaborative propose-and-evaluate agentic framework where the spatial and temporal reasoning are duly decoupled and re-organized in a collaborative manner between sub-agents during the retrieving process. 

As illustrated in~\Cref{fig:archi}, our framework consists of four key components: Controller, Spatial Reasoning Agent (SRA), Temporal Reasoning Agent (TRA), and a shared Candidate Memory: 

\begin{itemize}
    \item \textbf{Controller} oversees the whole pipeline, handles state control and, execution flow and processes input data for agents such as parsing the query and the video scenes. 
    \item \textbf{Spatial Reasoning Agent (SRA)} excels at spatial reasoning on single frames and is able to condition its output by referencing context information.
    \item \textbf{Temporal Reasoning Agent (TRA)} specializes in temporal observation and is capable for multiple sub-tasks such as providing verification regarding actions and temporal interactions, or performing temporal localization, depending on the instructions.
    \item \textbf{Candidate Memory} keeps track of the candidate tube masks that have been investigated by the Temporal Reasoning Agent. 
\end{itemize}

With these components, the retrieval process of the target entity is performed in a sequential order on given frames , where SRA proposes the candidate bounding box on the current frame; The Controller executes the tracking and adds its mask as visual prompts when passes to TRA, who serves an oracle peeking into the future, verifying and providing descriptive caption as context information, followed by temporal localization if it is the target. We will formulate and elaborate their roles and designs in the following subsections. We also provide an algorithm in~\Cref{alg:agentic_stvg}.

\subsection{Controller}

The Controller is the entry point of the agentic multi-modal retrieval process. It is responsible for query parsing, flow control and action execution.

\noindent \textbf{Parsing}. Queries for grounding related tasks can be in various forms, e.g., subject-relation-object triplet (Vidor~\cite{vidor}), declarative and interrogative sentences (VidSTG~\cite{stgrn}) and multiple sequential actions (HC-STVG~\cite{hcstvg}). Nevertheless, they can be parsed as $Q_{np}$ (subject noun phrase) and $Q_{context}$ (context) in general (except for the interrogative case). Widely used in image visual grounding tasks, $Q_{np}$ embodies the most important static cues; however, in video grounding, the critical cues are in $Q_{context}$ which are dynamic and temporal related descriptions. We invoke single modality LLM to parse the query into $Q_{np}$ and $Q_{context}$. For video parsing, we use PySceneDetect~\cite{pyscenedetect} to conveniently split the raw input clips into scene segments. As each scene will be processed independently, we will reuse the notation $V$ for each scene segment in this paper for brevity.

\noindent \textbf{Flow Control}. In the proposed agentic framework, the Controller acts as a lightweight orchestration module that facilitates interaction between SRA and TRA. It provides deterministic support functions, including entity tracking, visual prompting, and maintenance of the candidate memory $\mathcal{M}$ and global context $\mathcal{C}$ for inter-agent communication. All semantic judgments and decision-making are delegated to the agents; therefore, these support operations are described together with the corresponding agent behaviors in the following subsections.

\begin{algorithm}[!t]
\caption{Agentic Spatio-Temporal Grounding via Collaborative Reasoning}
\label{alg:agentic_stvg}
\begin{algorithmic}[1]
\Require Video $V=\{f_1,\dots,f_T\}$, query $Q$
\Ensure Trimmed tube $\hat{\mathcal{T}}$

\State Initialize memory $\mathcal{M}\gets\emptyset$, context $\mathcal{C}\gets\emptyset$
\State Initialize Controller state $\mathcal{S} \gets \textsc{Normal}$, time index $t\gets 1$
\State Define action space: \\
$\mathcal{A}= \{\textsc{Propose},\textsc{Track},\textsc{Verify},\textsc{Advance},\textsc{Fallback},\textsc{Terminate}\}$

\While{True}
    \State \textbf{OBSERVE:} $a \gets (f_t, \mathcal{S}, \mathcal{M}, \mathcal{C})$
    \State \textbf{DECIDE:} $a \gets \pi(o)$ \Comment{bounded policy}
    
\vspace{-0.6em}
\Statex \hspace{\algorithmicindent} \rule{0.9\linewidth}{0.4pt}
\vspace{-0.2em}
        
    \If{$a=\textsc{Propose}$}\Comment{SRA decision}
        \State $b_t \gets \textsc{MLLM}_{\text{SRA}}(f_t, Q, \mathcal{C})$
        \State $\mathcal{C}.\textsc{update}(b_t)$
        
\vspace{-0.6em}
\Statex \hspace{\algorithmicindent} \rule{0.9\linewidth}{0.4pt}
\vspace{-0.2em}

    \ElsIf{$a=\textsc{Track}$}\Comment{Controller execution}
        \State $\mathcal{T} \gets \textsc{Tracker}(V, b_t)$
        \If{$\mathcal{T}=\emptyset$}
            \State $\mathcal{C}.\textsc{update}(\text{``track-fail''})$
        \ElsIf{$\mathcal{T}\in \mathcal{M}$}
            \State $\mathcal{C}.\textsc{update}(\text{``duplicate''})$
        \Else
            \State $\mathcal{M}.\textsc{add}(\mathcal{T})$
            \State $V_{sp} \gets \textsc{SpatialPrompt}(V,\mathcal{T})$
        \EndIf
        
\vspace{-0.6em}
\Statex \hspace{\algorithmicindent} \rule{0.9\linewidth}{0.4pt}
\vspace{-0.2em}

    \ElsIf{$a=\textsc{Verify}$}\Comment{TRA Decision}
        \State $(\tau, c) \gets \textsc{MLLM}_{\text{TRA}}(V_{sp}, Q)$
        \If{$\tau \neq \emptyset$}
            \State $\hat{\mathcal{T}} \gets \textsc{Trim}(\mathcal{T}, \tau)$
            \State \textbf{DECIDE:} $a \gets \textsc{Terminate}$
        \Else
            \State $\mathcal{C}.\textsc{update}(c)$ %\Comment{context refinement}
        \EndIf
        
\vspace{-0.6em}
\Statex \hspace{\algorithmicindent} \rule{0.9\linewidth}{0.4pt}
\vspace{-0.2em}

    \ElsIf{$a=\textsc{Advance}$}\Comment{Controller execution}
        \State $t \gets t + \Delta$
        \If{$t > T$}
            \State $\mathcal{S} \gets \textsc{Fallback}$
        \EndIf
        
\vspace{-0.6em}
\Statex \hspace{\algorithmicindent} \rule{0.9\linewidth}{0.4pt}
\vspace{-0.2em}

    \ElsIf{$a=\textsc{Fallback}$}\Comment{Controller execution}
        \State $V_{tp} \gets \textsc{TemporalPrompt}(V)$
        \State $(\tau, c) \gets \textsc{MLLM}_{\text{TRA}}(V_{tp}, Q)$
        \State $V \gets \textsc{Trim}(V, \tau)$
        \State $\mathcal{M}\gets\emptyset$, $\mathcal{S} \gets \textsc{Normal}$, $\mathcal{C}\gets c$, $t\gets 1$
        
\vspace{-0.6em}
\Statex \hspace{\algorithmicindent} \rule{0.9\linewidth}{0.4pt}
\vspace{-0.2em}

    \ElsIf{$a=\textsc{Terminate}$}\Comment{Controller execution}
        \State \Return $\hat{\mathcal{T}}$
    \EndIf
\EndWhile
\end{algorithmic}
\end{algorithm}

\subsection{Spatial Reasoning Agent}

The main task of the SRA module is to propose all potential candidate object tubes with their spatial information given the query text. Intuitively this can be done with object detectors or image level visual grounding models; however, there are several challenges which make this naive solution coarse and inefficient for videos: \textbf{1)}. object detectors with closed or open vocabulary usually focus only on subject class labels or simple noun phrases mentioned in $Q_{np}$ while ignoring other information in $Q_{context}$; visual grounding models can capture spatial or relational cues but the temporal cues in $Q_{context}$ can cause confusions thus compromising their grounding accuracy; \textbf{2)}. the number of detected objects can be overwhelming in crowded scenes when $Q_{np}$ is too general, and even more redundant for video in a frame-by-frame routine where the same object is proposed repeatedly in similar frames; this poses a tremendous workload for TRA to verify and process.

To make the candidate generation process more efficient and flexible for video, we propose the spatial reasoning context $\mathcal{C}$ and Candidate Memory $\mathcal{M}$ to support SRA. Concretely, given the current frame $f_t$ from the sampled clip $V=\{f_1,\dots,f_T\}$, the parsed query $Q_{np}$ and $Q_{context}$, we engage MLLM with an emphasis on $Q_{np}$ but only taking $Q_{context}$ as a reference:
\begin{equation}
    \vect{b}_t = \textsc{MLLM}(f_t, Q_{np}, Q_{context}, \mathcal{C})
\end{equation}
The output bounding boxes $\vect{b}_t$ on current frame $f_t$ is then propagated efficiently through all frames in real-time via object tracking tools such as SAM2~\cite{sam2} to extract the candidate's spatio-temporal mask tube $\mathcal{T}$, which is then added to the Candidate Memory $\mathcal{M}$ if not already stored. 

The spatial reasoning context $\mathcal{C}$ is a global conversation memory maintained by the Controller, which keeps useful context and messages received from other agent modules via communication in the tube proposal generation process. This context memory can effectively condition the response of the MLLM with adaptive instructions in an collaborative manner, such as to consider other objects in the same frame, to skip the frame, or simply correct output formatting error depending on the context received. In this form, the SRA operates in a controlled multi-round dialogue session when propagating through the clip, effectively reducing the number of candidates and workload for TRA.

We highlight that the proposed SRA is different from traditional weakly-supervised or zero-shot methods where the proposal tube generation is performed offline and treated as a default pre-processing step; in contrast, the SRA utilizes context and memory to improve the robustness and efficiency, as well as the quality of the proposed tubes in an collaborative manner between agents.

\subsection{Temporal Reasoning Agent}
\label{subsec:fallback}

The TRA performs temporal reasoning over given frames which is a key requirement for video analysis. However, most existing video MLLMs~\cite{vdllm, vdllm_online, llavast, vdrefer} lack explicit spatial focus when performing the temporal reasoning process, which makes them prone to hallucinations or requires multiple rounds of interactive conversations to get the desired answer in the task of STVG; resulting in reliability and efficiency issues due to the massive input and output token throughput.

To make the temporal reasoning more robust and efficient as a component in an agentic system, we utilize both spatial and temporal visual prompts to handle the challenges. Concretely, the TRA undertakes four key roles in our proposed framework: \textbf{1)}. TRA as Scene Filter, 
\textbf{2)}. TRA as Candidate Tube Verifier, \textbf{3)}. TRA as Grounded Temporal Localizer, \textbf{4)}. TRA as Ungrounded Temporal Localizer.

\noindent \textbf{Scene Filter (SF)}. After the Controller parses the input clip into $m$ disjoint sub-scenes: $V = \bigcup_{i=1}^{m} V_i$ where $V_{i} \subseteq V$; the TRA directly judges the content of each scene without any form of prompt whether it is relevant to the query:
\begin{equation}
\begin{aligned}
    r_{dec} &= \textsc{MLLM}(V_{i}, Q)
\end{aligned}
\end{equation}
\noindent \textbf{Candidate Tube Verifier}. To assist the SRA in the tube candidate generation process, it is crucial for the temporal reasoning module to respond fast and reliably as the SRA is performed in a frame-by-frame fashion. Whenever the SRA confirms a new tube candidate, TRA is required to provide validation results as well as evidence on whether it is the target entity by reasoning over the video clip with a spatial focus on the target entity. Specifically, the Controller executes the spatial prompting process based on the tube mask $\mathcal{T}$ received from SRA, by putting visual markers (outlines and text) on the candidate~\cite{som} in video $V$ to produce the spatially prompted frames $V_{sp}$. The TRA is then prompted to force spatial focus on the marked entity, providing temporal related descriptive caption $r_{cap}$ only for the target entity:
\begin{equation}
\begin{aligned}
    V_{sp} &= \textsc{SpatialPrompt}(V, \mathcal{T})\\
    r_{dec}, r_{cap} &= \text{MLLM}(V_{sp}, Q_{np}, Q_{context})
\end{aligned}
\end{equation}
The verification process can be easily completed with an additional LLM inference $r_{dec} = \textsc{LLM}(r_{cap}, Q_{np}, Q_{context})$, but we opt to implement this in one single inference for efficiency consideration. In the case of unsuccessful validation results, $r_{cap}$ will be sent back to SRA as $context$ to guide its next prediction.

\noindent \textbf{Grounded Temporal Localizer}. Once the candidate tube $\mathcal{T}$ has been verified by the Tube Verifier, the Grounded Temporal Localizer is required to locate the time span that corresponds to the actions or contexts in $Q_{context}$. As we have altered the frame rate of the original video by sparse sampling, discrete time index is used instead of absolute time values. In this sub-task, the localizer requires a high level of temporal awareness as well as spatial focus; thus we apply temporal prompt visually as raw text (e.g. \texttt{"\#12"}) to explicitly indicate the underlining frame index visually on a fixed position of each frame in $V_{sp}$, resulting in a spatio-temporally prompted clip $V_{stp}$:
\begin{equation}
\begin{aligned}
    V_{stp} &= \textsc{TemporalPrompt}(V_{sp})\\
    \tau &= \textsc{MLLM}(V_{stp}, Q_{np}, Q_{context})
\end{aligned}
\end{equation}
where $\tau = (st,ed)$ is the predicted temporal span and the trimmed tube $\mathcal{\hat{T}} = \textsc{Trim}(\mathcal{T}, \tau)$ will be the final results for this sample.

\noindent \textbf{Ungrounded Temporal Localizer}. In case of fallback cases where the SRA has gone through all frames but still cannot find a valid tube, the TRA will initialize a "temporal localization first" strategy. Concretely, only temporal prompts are applied to the video $V$ without a candidate tube, the TRA has to reason temporally without spatial focus:
\begin{equation}
\begin{aligned}
    V_{tp} &= \textsc{TemporalPrompt}(V)\\
    \tau &= \textsc{MLLM}(V_{sp}, Q_{np}, Q_{context})
\end{aligned}
\end{equation}
The raw input is then trimmed as $V \gets \textsc{Trim}(V, \tau)$ for SRA and TRA to enter a second round of collaborative reasoning, with all state, memory and context cleared.

\begin{table*}[!t]
  \caption{Performance comparison with existing methods of different learning paradigms on the test set of VidSTG (Declarative Sentences) and HC-STVG (v1), validation set of HC-STVG (v2). }
  \label{tab:main_res}
  \centering
    % \begin{tabular}{p{3.0cm} c cccc c ccc c ccc}
    \begin{tabular}{p{3.0cm} c C{.9cm}C{.9cm}C{.9cm}C{.9cm} c ccc c ccc}
    % \hline
    \toprule
    \noalign{\vskip 1pt} 
    \multirow{2}{*}{\centering \textbf{Methods}} & &
    \multicolumn{4}{c}{\textbf{VidSTG (Declarative)}} & &
    \multicolumn{3}{c}{\textbf{HC-STVG (v1)}} & &
    \multicolumn{3}{c}{\textbf{HC-STVG (v2)}} \\
    \noalign{\vskip 2pt}
     & & \small m\_tIoU & \small m\_vIoU & \small vIoU@0.3 & \small vIoU@0.5 & & \small m\_vIoU & \small vIoU@0.3 & \small vIoU@0.5 & & \small m\_vIoU & \small vIoU@0.3 & \small vIoU@0.5\\
    \noalign{\vskip 4pt}
    \cline{1-1} \cline{3-6} \cline{8-10} \cline{12-14}
    \noalign{\vskip 4pt}
    
    \textbf{Fully-Supervised} & & & & & & & & & & & & \\ 
    \noalign{\vskip 2pt} 
    TubeDETR~\cite{tubedetr}~{\small\color{gray}CVPR'22} & & 48.1 & 30.4 & 42.5 & 28.2 & & 32.4 & \uline{49.8} & \uline{23.5} & & 36.4 & 58.8 & \uline{30.6} \\ 
    \noalign{\vskip 2pt} 
    STCAT~\cite{stcat}~{\small\color{gray}NIPS'22} & & 50.8 & 33.1 & 46.2 & 32.6 & & 35.0 & 57.7 & 30.0 & & \textemdash & \textemdash & \textemdash \\ 
    \noalign{\vskip 2pt} 
    CSDVL~\cite{csdvls}~{\small\color{gray}CVPR'23} & & \textemdash & 33.7 & 47.2 & 32.8 & & 36.9 & 62.2 & 34.8 & & 38.7 & 65.5 & 33.8 \\ 
    \noalign{\vskip 2pt} 
    CG-STVG~\cite{cgstvg}~{\small\color{gray}CVPR'24} & & 51.4 & 34.0 & 47.7 & 33.1 & & 38.4 & 61.5 & 36.3 & & 39.5 & 64.5 & 36.3 \\

    % \cline{1-1} \cline{3-6} \cline{8-10} \cline{12-14}
    % \hline
    \midrule
    \textbf{Weakly-Supervised} & &  & & &  & &  & &  & &  & & \\ 
    \noalign{\vskip 2pt} 
    % AWGU~\cite{awgu} & & \textemdash & 9.0 & 7.9 & 3.1 & & 8.2 & 4.5 & 0.8 & & \textemdash & \textemdash & \textemdash \\ 
    % Vis-Ctx~\cite{visctx}& & \textemdash & 9.3 & 7.3 & 3.3 & & 9.8 & 6.8 & 1.0 & & \textemdash & \textemdash & \textemdash \\ 
    WINNER~\cite{winner}~{\small\color{gray}CVPR'23} & & \textemdash & 11.6 & 14.1 & 7.4 & & 14.2 & 17.2 & 6.1 & & \textemdash & \textemdash & \textemdash \\
    \noalign{\vskip 2pt} 
    % CVTP*~\cite{cvtp} & & \textemdash & 17.9 & 22.4 & 14.9 & & 16.4 & 18.7 & 8.3 & & \textemdash & \textemdash & \textemdash \\
    VCMA~\cite{vcma}~{\small\color{gray}ECCV'24} & & \textemdash & 14.5 & 18.6 & 8.8 & & 14.6 & 18.6 & 5.8 & & \textemdash & \textemdash & \textemdash\\
    \noalign{\vskip 2pt} 
    CoSPaL~\cite{cospal}~{\small\color{gray}ICLR'25} & & 41.1 & 16.0 & 20.1 & 13.1 & & 22.1 & 31.8 & 19.6 & & 22.2 & 31.4 & 18.9\\
    \noalign{\vskip 2pt} 
    STPro~\cite{stpro}~{\small\color{gray}CVPR'25} & & \textemdash & 15.5 & 19.4 & 12.7 & & 17.6 & 27.0 & 12.9 & & 20.0 & 31.1 & 14.6 \\
    
    % \cline{1-1} \cline{3-6} \cline{8-10} \cline{12-14}
    \midrule
    \textbf{Zero-Shot} & &  & & &  & &  & &  & &  & & \\ 
    \noalign{\vskip 2pt} 
    RedCircle~\cite{red_circle}~{\small\color{gray}ICCV'23} & & \textemdash & 8.6 & 7.6 & 0.9 & & 9.1 & 7.8 & 1.6 & & \textemdash & \textemdash & \textemdash\\
    \noalign{\vskip 2pt} 
    ReCLIP~\cite{reclip}~{\small\color{gray}ACL'22} & & \textemdash & 19.1 & 29.4 & 10.6 & & 16.2 & 20.4 & 11.9 & & \textemdash & \textemdash & \textemdash\\
    \noalign{\vskip 2pt} 
    E3M~\cite{e3m}~{\small\color{gray}ECCV'24} & & \textemdash & 16.2 & 20.5 & 11.9 & & 19.1 & 29.4 & 10.6 & & \textemdash & \textemdash & \textemdash\\
    \noalign{\vskip 2pt} 
    LN-STVG~\cite{lnstvg}~{\small\color{gray}NIPS'25} & & \textemdash & 18.0 & 29.8 & 12.2 & & 24.8 & 41.5 & 16.3 & & 27.7 & 44.7 & 19.5 \\
    \noalign{\vskip 2pt} 
    % RealVG~\cite{realvg}~{\small\color{gray}ACMM'25} & & 37.4 & 29.0 & 35.0 & 25.5 & & \textbf{29.5} & 40.0 & \textbf{25.8} & & 33.1 & 42.3 & 27.0 \\
    
    RealVG~\cite{realvg}~{\small\color{gray}ACMM'25} & & 37.4 & 29.0 & 35.0 & 25.5 & & 29.5 & 40.0 & 25.8 & & 33.1 & 42.3 & 27.0 \\
    
    \midrule
    % \cline{1-1} \cline{3-6} \cline{8-10} \cline{12-14}
    % ASTG~{\small\color{gray} QW2.5-VL-72B} & & 45.6 & 29.2 & 40.3 & 27.8 & & - & - & - & & - & - & - \\
    % ASTG~{\small\color{gray} QW3-VL} & & - & - & - & - & & 28.3 & 47.7 & 25.3 & & 34.8 & 57.3 & 31.4 \\
    %combine
    % ASTG~{\small\color{gray} QW3-VL-plus} & & \textbf{45.6} & \textbf{29.2} & \textbf{40.3} & \textbf{27.8} & & 28.3 & \textbf{47.7} & 25.3 & & \textbf{34.8} & \textbf{57.3} & \textbf{31.4} \\
    
    \noalign{\vskip 1pt} 
    ASTG (Ours) & & \textbf{45.6} & \textbf{29.2} & \textbf{40.3} & \textbf{27.8} & & \textbf{32.3} & \textbf{\uline{54.4}} & \textbf{\uline{28.2}} & & \textbf{34.8} & \textbf{55.5} & \textbf{\uline{31.4}} \\
    \noalign{\vskip 1pt} 
    
    \bottomrule
  \end{tabular}
\end{table*}

\section{Experiments}

\subsection{Datasets and Evaluation Metrics}
We conduct the experiments on three widely adopted video grounding benchmarks: VidSTG~\cite{stgrn}, HC-STVG v1 and HC-STVG v2~\cite{hcstvg}. 

\noindent\textbf{VidSTG} dataset is developed based on Vidor~\cite{vidor} where the labels are always in a form of $<subject, relation, object>$ triplet. By rephrasing of the triplet into declarative and interrogative sentences, VidSTG is constructed with $99{,}943$ sentence-video pairs in total and $10{,}303$ samples for the testing set ($4{,}610$ for declarative sentences and $5{,}693$ for interrogative sentences). The duration of the video samples span from $3$ seconds to $2$ minutes with an average length of $28.01$ seconds and the average length of the query sentences are $11.12$ and $8.98$ words for declarative and interrogative cases, respectively. The object class label targeted in VidSTG are mainly covered by the same categories in MS-COCO~\cite{coco}. 

\noindent\textbf{HC-STVG}~\cite{hcstvg} is constructed in two versions: v1 and v2; which only focus on human subject. The emphasis HC-STVG is mainly on temporal action sequences, and there are roughly $2.5$ actions or interactions mentioned on average in a sentence query. The duration of each video is $20$ seconds. There are $5{,}660$ video-query pairs in total with $1{,}160$ samples in test set for v1; and $16{,}685$ samples are split into $10{,}131$ training, $2{,}000$ validation and $4{,}413$ test samples for v2. Since no annotation is provided for the test set of HC-STVG v2, the validation set is widely used for evaluation. 

\noindent\textbf{Evaluation Metrics}. Following previous works, $\mathbf{m\_tIoU}$, $\mathbf{m\_vIoU}$ and $\mathbf{vIoU@R}$ are used for evaluation and comparison. $\mathbf{tIoU}$ is calculated between ground truth and predicted time span. The per sample $\mathbf{vIoU}$ is defined by $\frac{1}{|\mathcal{F}_U|} \sum_{t \in \mathcal{F}_I}{\rm IoU}(\bm{b}^t_{gt}, \bm{b}^t_{pred})$, where $\mathcal{F}_I$ and $\mathcal{F}_U$ indicates the set of intersection and union of the ground truth and predicted frames, respectively. Lastly, $\mathbf{vIoU@R}$ measures the percentage of samples whose $\mathbf{vIoU}$ is above the threshold $R$.

\subsection{Implementation Details}
In this work, we mainly adopt Qwen3~\cite{qwen3} series, including Qwen-plus and Qwen3-VL-plus, as our LLM and MLLM for SRA, TRA and the query parsing job of the Controller. We utilize PySceneDetect~\cite{pyscenedetect} and TRA to parse and filter the irrelevant scene before starting the main pipeline. We incorporate SAM2~\cite{sam2} into our framework as a tool to track and extract the candidate tube given its bounding box on one specific frame, which is obtained from response of SRA. For efficiency considerations: the frame resolution is set to $448$ for both SRA and SAM2, and $336$ for TRA; the video clips are sampled at a frame rate of $3$ and $2$ per second for VidSTG and HC-STVG datasets, respectively. The memory module $\mathcal{M}$ is implemented to store the tracked masks of distinct entities on all frames. New candidate tube is confirmed if its mask $IoU$ with all stored tubes is less than a threshold, which is set to $0.5$. We add visual and temporal prompt following \cite{som} and \cite{numberit}.

\begin{table*}[!t]
  \caption{Ablation experiment of core module choices on the test set of HC-STVG v1. Different module on each row is added sequentially.}
  \label{tab:ab1}
  \centering
    \begin{tabular}{p{7.0cm} c cccc c c}
    % \hline
    \toprule
    \noalign{\vskip 1pt} 
    \multirow{2}{*}{ \textbf{Modules} } & &
    \multicolumn{4}{c}{\textbf{Metrics}} & & \multirow{2}{*}{ \textbf{Latency (s)} }\\
    \noalign{\vskip 2pt}
     & & m\_tIoU & m\_vIoU & vIoU@0.3 & vIoU@0.5 & & \\
     
    \noalign{\vskip 2pt}
    \cline{1-1} \cline{3-6} \cline{8-8} 
    \noalign{\vskip 3pt}
     
    Baseline (SRA $+$ Tube Verifier) & & 28.8 & 19.0 & 26.1 & 11.0 & & 23.9  \\ 
    \noalign{\vskip 2pt} 
    
    $+$ Scene Filter & & 35.1 & 22.4 & 32.5 & 16.2 & & 24.2  \\ 
    \noalign{\vskip 2pt} 
    
    \hspace*{1em} $+$ Grounded Temporal Localizer & & 36.6 & 25.8 & 36.1 & 19.2 & & 28.2  \\
    \noalign{\vskip 2pt}  
    \hspace*{1em} $+$ Grounded Temporal Localizer (Thinking) & & 43.8 & 28.3 & 45.4 & 24.6 & & 71.7  \\
    \noalign{\vskip 2pt}  
    
    \hspace*{2em} $+$ Ungrounded Temporal Localizer & & 44.2 & 29.1 & 46.7 & 25.2 & & 77.3  \\ 
    \noalign{\vskip 2pt}  
    \hspace*{2em} $+$ Ungrounded Temporal Localizer (Thinking) & & 47.7 & 32.3 & 54.4 & 28.2 & & 94.8  \\ 

    \bottomrule
  \end{tabular}
\end{table*}

\subsection{Quantitative Evaluation}
We compare our main results with existing SOTA methods on the three datasets from multiple learning paradigms including weakly-supervised and zero-shot settings in~\Cref{tab:main_res}. We also list the results of the SOTA fully-supervised methods for reference. Notably, the performance of both fully and weakly supervised methods are attained by utilizing partial or full data in the training corpus; thus the trained models are sensitive to data distribution shift and their performance usually drops significantly when applied to open-world scenario in a zero-shot setting, and this can only be alleviated by re-training or fine-tuning on new data. On the other hand, the zero-shot methods are more robust to data distribution shift and are able to leverage the vast knowledge base stored in foundation models or MLLMs.

Specifically, RedCircle~\cite{red_circle}, ReCLIP~\cite{reclip} and E3M~\cite{e3m} are constructed based on the foundation model CLIP~\cite{clip}. LN-STVG~\cite{lnstvg} and RealVG~\cite{realvg} leverage MLLMs in their pipelines, however, both works follow a deterministic workflow and MLLMs are used as tools or embedding extractors instead of agents that can adapt their answers according to the uncertain environment. Our proposed agentic method has outperformed all existing weakly and zero-shot methods consistently. For $\mathbf{vIoU@0.3}$, a significant improvement of $4.7\,\%$, $14.4\,\%$ and $13.2\,\%$ over RealVG~\cite{realvg} is observed on the three datasets, respectively. And for $\mathbf{vIoU@0.5}$, the gain is $2.3\,\%$, $2.4\,\%$ and $4.4\,\%$. This indicates the candidate tubes retrieved by our framework have a notable quality improvement in terms of spatio-temporal precision. Additionally, our method achieves a substantial gain of $8.2\,\%$ for $\mathbf{tIoU}$ on the VidSTG dataset (declarative).

We notice that the performance of our proposed agentic system has been comparable to one of the fully-supervised baseline methods TubeDETR~\cite{tubedetr} in all evaluation metrics. The proposed method even achieves better performance for $\mathbf{vIoU@0.5}$ on HC-STVG, where we underline the numbers in~\Cref{tab:main_res}.

\begin{table}
  \caption{Ablation experiment (with baseline) on the skimming stride of the SRA on the input frames. }
  \label{tab:ab2}
  \centering
    \begin{tabular}{C{1.0cm} c C{.9cm}C{.9cm}C{.9cm}C{.9cm} c C{.9cm}}
    % \hline
    \toprule
    \noalign{\vskip 1pt} 
    \multirow{2}{*}{ \textbf{Stride} } & &
    \multicolumn{4}{c}{\textbf{HC-STVG (v1)}} & & \multirow{2}{*}{\small\textbf{Latency (s)}}\\
    \noalign{\vskip 2pt}
     & & \small m\_tIoU & \small m\_vIoU & \small vIoU@0.3 & \small vIoU@0.5 &  \\
    \noalign{\vskip 2pt}
    \cline{1-1} \cline{3-6} \cline{8-8}
    \noalign{\vskip 3pt}
     
    $\Delta = 1$ & & 29.2 & 19.4 & 26.7 & 12.0 & & 41.2  \\ 
    \noalign{\vskip 2pt} 
    $\Delta = 2$ & & 28.8 & 19.0 & 26.1 & 11.0 & & 23.9  \\ 
    \noalign{\vskip 2pt} 
    $\Delta = 3$ & & 26.7 & 17.6 & 23.5 & 10.2 & & 16.5  \\ 
    \noalign{\vskip 2pt} 
    
    \bottomrule
  \end{tabular}
\end{table}

\subsection{Ablation Experiments}

% \begin{table*}[!htbp]
% \centering
% \caption{Ablation on prompts used on HC-STVG dataset.}
% \label{tab:ab2}
% % \begin{tabular}{p{.15cm}p{.15cm}p{.15cm}p{.01cm}p{.95cm}p{.95cm}p{.95cm}p{.95cm}p{.95cm}p{.95cm}}

% % \begin{tabular}{p{.25cm}p{.25cm}p{.25cm} c ccccc}
% \begin{tabular}{cc c ccccc}
% \toprule
% % \noalign{\vskip 2pt}
% \multicolumn{2}{c}{\small \textbf{Enable Thinking}}& &
% \multicolumn{5}{c}{\centering\textbf{Metrics}}\\ 
%  SF &  GTL \&  UTL & & m\_tIoU & m\_vIoU &  vIoU@0.3 &  IoU@0.5 &  Latency (s)\\
% \noalign{\vskip 2pt}
% \cline{1-2} \cline{4-8}
% \noalign{\vskip 4pt}

%  % & \textit{Skip} & & 7.13 & 5.87 & 1.73 & 18.64 & 21.5\\
% % \noalign{\vskip 2pt}

%  & & & 7.13 & 5.87 & 1.73 & 18.64 & 26.7\\
% \noalign{\vskip 2pt}

% \checkmark & & & 13.12 & 14.11 & 5.28 & 41.77 & 32.4\\
% \noalign{\vskip 2pt}

% & \checkmark & & 14.02 & 16.32 & 6.39 & 44.73 & 73.2 \\
% \noalign{\vskip 2pt}

% \checkmark & \checkmark & & 48.1 & 31.3 & 47.7 & 28.3 & 44.73\\
% \bottomrule
% \end{tabular}
% % }
% \end{table*}

% sloc on v2, 24s
% tloc on v2, 50s

We conduct ablation experiments on the core modules of the proposed method in~\Cref{tab:ab1}. Since our framework follows an object centric design and Propose-and-Evaluate methodology, the SRA and the Tube Verifier (TRA) will be the essential modules which form the baseline. As shown in~\Cref{tab:ab1}, the performance of the baseline on HC-STVG v1 has been comparable to the SOTA weakly-supervised methods STPro~\cite{stpro} listed in ~\Cref{tab:main_res}. 

We observe that in ~\Cref{tab:ab1}, the Scene Filter contributes notably to the $tIoU$ metric. We note that this is because the HC-STVG benchmark is collected from movie scenes which contain a lot of scene transitions where correctly filtering irrelevant scenes is especially crucial to our framework because the tracker usually fails in multi-scene scenario thus leads to unsuccessful retrieval due to frequent tool usage failures. Intuitively, the Scene Filter as a pre-processing step would incur additional latency, but in our framework the overall overhead is minor as the module also effectively reduces the number of frames for the subsequent modules to analyze. 

In addition, we also examine the choices on the working mode of TRA's core modules including the Grounded and Ungrounded Temporal Localizers. The Candidate Tube Verifier is set to be non-thinking mode only due to efficiency considerations. It can be observed in~\Cref{tab:ab1} that by activating the thinking mode of the TRA for both Grounded and Ungrounded Temporal Localizer, the overall performance is improved by a significant margin at the cost of a high inference latency (~$43$ seconds on average); while the non-thinking localizers are fast but only provides marginal gains. We find this is one of the inherent limitations modern reasoning MLLMs share when dealing with sequences. In our case, the MLLMs frequently hallucinate about the frame index of the given frames, even when it is explicitly numbered and visually prompted. It is observed in our experiments that the thinking mode can alleviate hallucinations when the model keeps correcting itself at the cost of an extended inference latency and token consumption.

We study the impact of skimming stride $\Delta$ of the SRA in~\Cref{tab:ab2}. Aside from the frame sampling rate which is fixed in all our experiments, the skimming stride $\Delta$ also controls the granularity of the temporal observation. However, in our framework, the TRA only needs to infer once per tube while the SRA needs to skim through the whole sequence. Therefore, we only conduct this experiment on the baseline models, i.e. only with the SRA and the Tube Verifier modules because the Temporal Localizer is not affected by $\Delta$. It is a trade-off for the SRA: a higher stride increases inference efficiency but risk missing the correct target; while a lower stride introduces redundant workload and higher latency in exchange for higher candidates coverages. We determine the best stride $\Delta=2$ according to~\Cref{tab:ab2} because only marginal gains are observed with a lower stride at the cost of almost doubling the latency, while the higher stride results in a notable performance drop.

\begin{figure*}[!t]
  \centering
  \includegraphics[width=\textwidth]{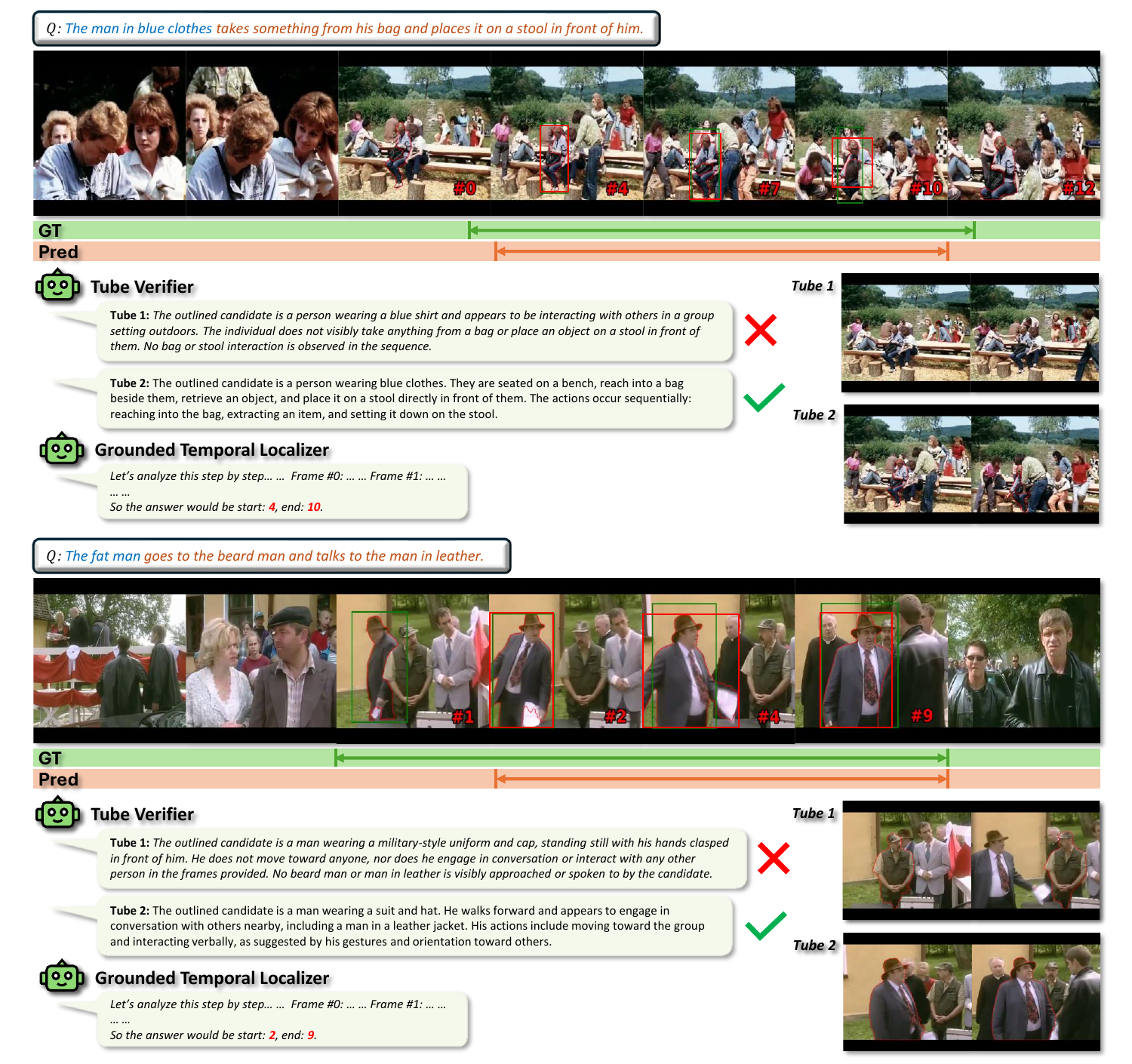}
  \caption{Qualitative examples of the proposed ASTG method on HC-STVG v2.}
  \label{fig:qeg}
\end{figure*}

\subsection{Qualitative Examples}
% \label{subsec:qeg}
We provide visualizations of the intermediate and final prediction results in~\Cref{fig:qeg}, compared with the ground truth bounding boxes and temporal boundaries. It can be illustrated in~\Cref{fig:qeg} that the mask visual prompts added to raw frames can effectively help MLLM to focus on concrete concepts (objects and activities) thus make sound judgements and reduce hallucinations in a complex and ambiguous spatial-temporal scene. By utilizing the strong reasoning capabilities of the MLLMs, we design COT prompts together with explicit temporal markers to improve the robustness of the self-governed temporal localization process. We note that our method retrieve candidate mask tubes instead of bounding box tubes, which provides a more accurate spatial localization capability. In practice, we convert the mask tube into bounding box tube with morphological operations for quantitative evaluations.

\subsection{Conclusion}

In this paper, we propose ASTG, a training-free agentic spatio-temporal video grounder to address the challenges where expensive training data and poor generalization ability limit the application of traditional grounding systems in an open-world scenario. To better leverage the potential of the MLLMs, we introduce a novel agentic collaborative framework with multiple visual agents. Leveraging explicit spatial and temporal prompts, the proposed method duly decouples the spatial and temporal grounding process and re-organize them into an object centric propose-and-evaluate workflow with autonomy. We also introduce a visual memory and a textual dialogue as context to improve the inference efficiency as well as guiding the agent's behavior adaptively. The performance achieved has been comparable to some of the supervised methods, we hope this work can provide inspirations to more real-world STVG applications in a training-free or zero-shot scenario.


% \begin{acks}
% \end{acks}

\bibliographystyle{ACM-Reference-Format}
\bibliography{ref}


% \appendix









\end{document}